\documentclass[twoside]{article}

\usepackage[accepted]{aistats2025}
\usepackage{graphicx} 
\usepackage{amsmath}
\makeatletter

\usepackage{url}
\usepackage{csquotes}
\usepackage{nicefrac}
\usepackage{amsmath,amsthm,amssymb}
\usepackage{graphicx}
\usepackage{algorithm}
\usepackage{multirow}
\usepackage[utf8]{inputenc}
\usepackage{bbm}
\usepackage{hyperref}
\usepackage[dvipsnames]{xcolor}
\usepackage{tabularx}
\usepackage{booktabs,multirow,multicol}
\usepackage{soul}
\usepackage{tikz}
\usetikzlibrary{bayesnet}
\usetikzlibrary{arrows}

\def\PP{{\mathbb P}}     
\def\EE{{\mathbb E}}
\def\PP{{\mathbb P}}
\def\11{{\mathbf 1}}    

\def\cA{{\mathcal A}}        \def\cN{{\mathcal N}}      \def\cU{{\mathcal U}} \def\cD{{\mathcal D}}   \def\cP{{\mathcal P}}  \def\cE{{\mathcal E}}   \def\cQ{{\mathcal Q}}      \def\cX{{\mathcal X}}   

\def\ba{{\mathbf a}} \def\bb{{\mathbf b}}         \def\bk{{\mathbf k}}         \def\bt{{\mathbf t}}    \def\bx{{\mathbf x}} \def\by{{\mathbf y}}  

\def\bA{{\mathbf A}}          \def\bK{{\mathbf K}}  \def\bM{{\mathbf M}}           \def\bX{{\mathbf X}}   

\def\bf{{\mathbf f}} 
 
\def\fhigh{\bar{f}}
\def\flow{\ubar{f}}

\def\bk{\boldsymbol{k}}

\DeclareSymbolFont{boldoperators}{OT1}{cmr}{bx}{n}
\SetSymbolFont{boldoperators}{bold}{OT1}{cmr}{bx}{n}

\newcommand{\setbrac}[1]{\left\{{#1}\right\}}
\newcommand{\circbrac}[1]{\left({#1}\right)}
\newcommand{\squarebrac}[1]{\left[{#1}\right]}

\newcommand{\norm}[1]{\left\lvert #1 \right\rvert}

\theoremstyle{plain}
	\newtheorem{theorem}{Theorem}[section] 
	\newtheorem{proposition}[theorem]{Proposition}        
	\newtheorem{corollary}[theorem]{Corollary}

\theoremstyle{definition}
	\newtheorem{definition}[theorem]{Definition}

 \newtheorem{assumption}[theorem]{Assumption}

\theoremstyle{remark}

\theoremstyle{plain}
    \newtheorem{innercustomgeneric}{\customgenericname}
    \providecommand{\customgenericname}{}
    \newcommand{\newcustomtheorem}[2]{%
      \newenvironment{#1}[1]
      {%
       \renewcommand\customgenericname{#2}%
       \renewcommand\theinnercustomgeneric{##1}%
       \innercustomgeneric
      }
      {\endinnercustomgeneric}
    }
    
    \newcustomtheorem{customthm}{Theorem}
    \newcustomtheorem{customlemma}{Lemma}
    \newcustomtheorem{customprop}{Proposition}
    \newcustomtheorem{customcor}{Corollary}
\usepackage{algpseudocode}
\usepackage{caption}
\usepackage{subcaption}
\usepackage{todonotes}
\usepackage{mathtools}
\colorlet{standard_edge_color}{black}
\colorlet{standard_node_color}{white}
\everypar{\looseness=-1}

\usepackage[symbol]{footmisc}

\colorlet{varying_edge_color}{red!80!black}
\colorlet{varying_node_color}{red!10}
\usetikzlibrary{automata,arrows,calc,positioning,shapes}
\usepackage{lipsum}

\usetikzlibrary{arrows.meta}
\usetikzlibrary{bayesnet}
\usetikzlibrary{arrows}
\usepackage{accents}
\DeclareRobustCommand{\ubar}[1]{\underaccent{\bar}{#1}}
\newlength{\verticalsep}
\newlength{\horizontalsep}
\usepackage{multirow}

\usepackage[round]{natbib}

\begin{document}

%

%

\twocolumn[

\aistatstitle{The Hardness of Validating Observational Studies with Experimental Data}

\aistatsauthor{ Jake Fawkes\footnotemark[1]}
\aistatsaddress{ University of Oxford}

\aistatsauthor{ Michael O'Riordan \And  Athanasios Vlontzos \And Oriol Corcoll \And Ciar\'{a}n M. Gilligan-Lee }

\runningauthor{Jake Fawkes, Michael O'Riordan,  Athanasios Vlontzos, Oriol Corcoll, Ciar\'{a}n M. Gilligan-Lee }

\aistatsaddress{   Spotify \And  Spotify \& \\ Imperial College London  \And Spotify \And Spotify \& \\ University College London} ]

\footnotetext[1]{Work done during internship at Spotify.  Correspondence to: \texttt{jake.fawkes@st-hughs.ox.ac.uk}.}
\begin{abstract}
\vspace{-1em}

Observational data is often readily available in large quantities, but can lead to biased causal effect estimates due to the presence of unobserved confounding. 
Recent works attempt to remove this bias by supplementing observational data with experimental data, which, when available, is typically on a smaller scale due to the time and cost involved in running a randomised controlled trial. 
In this work, we prove a theorem that places fundamental limits on this ``best of both worlds'' approach. 
Using the framework of impossible inference, we show that although it is possible to use experimental data to \emph{falsify} causal effect estimates from observational data, in general it is not possible to \emph{validate} such estimates. 
Our theorem proves that while experimental data can be used to detect bias in observational studies, without additional assumptions on the smoothness of the correction function, it can not be used to remove it.
We provide a practical example of such an assumption, developing a novel Gaussian Process approach to construct intervals which contain the true treatment effect with high probability, both inside and outside of the support of the experimental data.  
We demonstrate our methodology on both simulated and semi-synthetic datasets and make the \href{https://github.com/Jakefawkes/Obs_and_exp_data}{code available}.
\end{abstract}

\section{INTRODUCTION}
\vspace{-1em}
It is often said that randomised controlled trials (RCTs) are the gold standard for establishing causal relationships \citep{hariton2018randomised,gilligan2022leveraging}, and estimating treatment effects \citep{aronow2021nonparametric, gilligan2020causing}. However, in many cases, it is prohibitively costly, slow, or even unethical to run experiments that are large enough to accurately estimate such effects. Meanwhile, observational data is abundant, being significantly easier and cheaper to obtain. Unfortunately, such data is often subject to unmeasured confounding. This leaves treatment effects unidentifiable, and naively attempting to use the observational data despite this will lead to biased estimates of causal effects. 

As a response to these problems, there has been substantial recent research effort to develop methodology for combining experimental and observational data sources, aiming to get the strengths of each \citep{colnet2024causal,lin2023many,van2023estimating,jeunen2022disentangling}. These approaches aim to use the experimental data to de-bias the observational data \citep{yang2020improved,kallus2018removing}, falsify observational studies \cite{hussain2022falsification,hussain2023falsification}, or benchmark the level of unmeasured confounding \citep{de2024detecting,de2024hidden}.

In this work, we prove a selection of fundamental limitations of this approach. We use the framework of impossible inference \citep{bahadur1956nonexistence,bertanha2020impossible}, a popular tool in econometrics \citep{canay2013testability}. This field studies when it is possible for a non-trivial hypothesis tests to exist for a problem, where trivial hypothesis tests are those which are unable to distinguish \emph{any} alternative from the null.
We apply this to show that whilst non-trivial tests exist to \emph{falsify} estimates from observational studies, we cannot \emph{validate} heterogeneous treatment effects estimates using experimental data without additional assumptions. In terms of benchmarking confounding with a causal sensitivity model, our result corresponds to the statement that it is possible to form valid \emph{lower bounds} of the sensitivity parameter but that it is impossible to form non-trivial upper bounds---again, absent additional assumptions. 



Our hardness proof relies on the lack of smoothness in the unknown correction function. Therefore, as an example of an assumption that does permit this type of inference, we take the corrective function to be a sample from a Gaussian Process, a probabilistic function family with inherent smoothness guarantees. Developing this into a workable and practical methodology leads us to create a novel Gaussian Process based approach to learning from pseudo-outcomes \citep{kennedy2023towards}, which correctly accounts for the unwieldy error distribution of pseudo-outcomes. We experimentally evaluate our approach against other Gaussian Process effect estimation approaches, showing strong improvement in predictive performance and uncertainty calibration. 
In short, to summarise our contributions: 
\begin{itemize}
    \item A proof of the limits of current approaches that use a learned corrective term to reconcile experimental and observational data. 
     \item A demonstration that the smoothness properties of Gaussian Processes circumvent the assumptions required for the above proof and provide guarantees that Gaussian Processes give intervals which contain the treatment effect over the whole observational support with high probability.
    \item A novel Gaussian Process method to learn from inverse propensity weighted pseudo-outcomes---which may be of independent interest.
    \item  Extensive experimentation validating the aforementioned novel method on synthetic and semi-synthetic data.
\end{itemize}
\section{BACKGROUND AND NOTATION}

\subsection{Notation}

We let the random variables $X$, $T$, and $Y$ represent the covariates, treatment, and outcomes, with domains $\cX$, $\{0,1\}$, and $\mathbb{R}$ respectively, and use $\bx$, $t$, and $y$ to denote realisations of the variables. We suppose we have two datasets, the observational $\cD_{e} = \setbrac{\bx^o_i,t^o_i,y^o_i}^{n_{o}}_{i=1}$ and the experimental $\cD_{o} = \setbrac{\bx^e_i,t^e_i,y^e_i}^{n_{e}}_{i=1}$ which are drawn from distributions $P_{e}(\bx,t,y)$ and $P_{o}(\bx,t,y)$ respectively, with $\cD = \cD_{o} \cup \cD_{e}$ denoting the full dataset of size $n=n_o+n_e$. We will use a variable $E$ to denote if we are in the observational or experimental regime, so that, for example, $P_{e}(\bx,t,y) = P(\bx,t,y \mid E=e)$. Letting $\star \in \setbrac{o,e}$ we use $\cX^{\star} \subset \cX$ for the support of $P_{\star}(\bx)$ and assume that $\cX^{e} \subset \cX^{o}$. Vectors of observations and treatment in a dataset, $\cD_{\star}$, are denoted by $\by_{\star}$ and $\bt_{\star}$ respectively, while $\bX_{\star}$ refers to the data matrix, so that $\by_{\star} = (y_i)^{n_e}_{i=1}$, $\bt_{\star} = (t_i)^{n_e}_{i=1}$, and $\bX_{\star} = (\bx_i)^{n_e}_{i=1}$. We let $\omega_{\star}(\bx)$ be defined as:
\begin{align}
    \omega_{\star}(\bx) &= \EE[Y|X=\bx,T=1,E=\star] \\
    &- \EE[Y|X=\bx,T=0,E=\star],
\end{align}
be the difference between expected conditional outcomes for each treatment conditional on $E=\star$ and $X=\bx$. We use potential outcomes \citep{rubin1974estimating}, so that $Y(t)$ represents the outcome from setting $T=t$\footnote{We make use of the SWIG framework to combine causal graphical models with potential outcomes. More details can be found in \citet{richardson2013single}.}.

\subsection{Objectives and Assumptions}

Throughout we focus on estimating the \textit{Conditional Average Treatment Effect} (CATE) for the observational datasets, which is given by:
\begin{align}
    \tau(\bx) \coloneqq \EE \left[Y(1) -Y(0) | X=\bx,E=o \right].
\end{align}
\begin{figure}
    \centering
\begin{tikzpicture}
    \node (T) [state, standard_edge_color, fill=standard_node_color] {$T$};
    \node (mid) [state, standard_edge_color, fill=standard_node_color, right=\horizontalsep of T,draw=none] {};
    \node (X) [state, standard_edge_color, fill=standard_node_color, above=\verticalsep of mid] {$X$};
    \node (Y) [state, standard_edge_color, fill=standard_node_color, right= \horizontalsep of mid] {$Y$};
    \node (U) [state, standard_edge_color, fill=standard_node_color, above= 0.5*\verticalsep of X] {$U$};
    \draw[->, thick, standard_edge_color] (T) edge (Y);
    \draw[->, thick, standard_edge_color] (X) edge (Y);
    \draw[->, thick, standard_edge_color] (X) edge (T);
    \draw[->, dashed, standard_edge_color,in =90,out = 180] (U) edge (T);
    \draw[->, standard_edge_color,in= 90, out = 0] (U) edge (Y);
    \draw[->, standard_edge_color] (U) edge (X);
    \end{tikzpicture}
    \caption{Causal Structure for generating the experimental and observational datasets. Dashed edges are only present in the observational dataset, whilst all others are present and fixed across both datasets.}
    \label{fig:generating_DAG}
\end{figure}
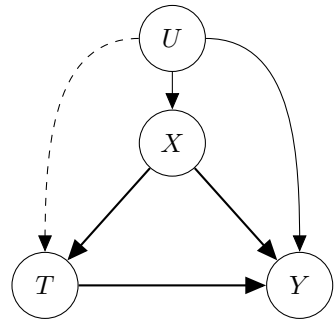
We assume the datasets are generated according to the causal structure in Figure \ref{fig:generating_DAG}, where dashed edges are present only in the observational dataset. Importantly, this implies $Y(t) \perp E \mid X= \bx$, which ensures that the CATE is fixed across environments as:
\begin{align}
    \tau(\bx) &= \EE \left[Y(1) -Y(0) | X=\bx,E=o \right] \\
    &= \EE \left[Y(1) -Y(0) | X=\bx,E=e \right].
\end{align}
We demonstrate this in Appendix \ref{ap:constant_CATE}.

For the experimental study we assume that treatment is randomised according to a known propensity score $\pi(\bx) = P_e(T=1 \mid X=x)$ which we assume to satisfy \textit{strict overlap} \citep{d2021overlap}. That is we assume that there exists a $\delta > 0$ such that:
\begin{align}
    \delta < \pi(\bx) < 1-\delta \text{ for all } \bx \in \cX^{e}.
\end{align}
Under the additional assumption of consistency ($Y(t) = Y$ when $T=t$) we have that the CATE is \textit{identified} \citep{pearl2009causal,richardson2013single} within the support of the experimental dataset, and given by:
\begin{align}
    \tau(\bx) = \omega_{e}(\bx) \text{ for all } \bx \in \cX^{e}.
\end{align}
This is not the case in the observational dataset, where the hidden confounding induced by $U$ means that $\tau(\bx) \neq \omega_{o}(\bx)$ in general. We use $\Delta(\bx)$ to denote this gap as:
\begin{align}
     \Delta(\bx) = \tau(\bx) - \omega_{o}(\bx). 
\end{align}
So that if $\Delta(\bx) = 0$, an unbiased estimate of $\omega_{o}(\bx)$ from the observational study is an unbiased estimate of the true CATE. Throughout, we will assume a model $\hat{\omega}_o(\bx)$ has been fit for ${\omega}_o(\bx)$ from the observational sample and let $\hat{\Delta}(\bx) \coloneqq \tau(\bx) - \hat{\omega_{o}}(\bx)$.

The idea is that $\Delta(\bx)$ should be simpler than the true CATE function, $\tau(\bx)$. Under such circumstances, it should be more efficient to use to the experimental dataset to estimate or bound $\hat{\Delta}(\bx)$ and combine it with $\hat{\omega_{o}}(\bx)$ than use the small experimental dataset to learn the CATE directly\citep{yang2020improved}. Ideally, we would want these to be extendable from the support of the experimental distribution to the support of the observational distribution \citep{kallus2018removing}, potentially by incorporating bounds on the correction function as opposed to point estimates. This would give us expression for CATE over all of $\cX^o$.

Finally, we introduce the IPW pseudo-outcome \citep{kennedy2023towards,curth2021nonparametric}, which throughout we will only refer to relative to the experimental distribution, as follows:

\begin{definition}[IPW Pseudo-Outcome]
  The IPW Pseudo-Outcome is given by:
\begin{align}
    \Tilde{Y} \coloneqq \circbrac{\frac{T-\pi(X)}{\pi(X)\circbrac{1-\pi(X)}}} Y
\end{align}
Where $\pi(X) = P(T=1\mid X,E=e)$. $\Tilde{Y}$ has the property that $\EE[\Tilde{Y} |X=\bx, E=e] = \tau(\bx)$.
\end{definition}
We assume that $\pi(X)$ is known, which is common for a well conducted randomised controlled trial. 
\subsection{Related work bounding $\Delta(\bx)$}

Our setting is first considered in \citet{kallus2018removing}, where they assume that $\Delta(\bx)$ is linear in order to allow for extrapolation from the experimental sample. By taking $\Delta(\bx)= \beta_0^{\top} \bx$ and assuming $\beta_0$ is identifiable from experimental data, they can obtain an estimate  $\Delta(\bx)$ that generalises beyond the experimental sample by performing a linear regression of $\{\bx^e_i\}^{n_e}_{i=1}$ onto $\{ \Tilde{y}_i^e-\hat{\omega}_o(\bx)\}^{n_e}_{i=1}$. \citet{kallus2018removing} prove that as the number of experimental and observational samples tend to infinity this will converge to the true CATE at a faster rate than using the experimental sample alone. This has been extended to the semiparametric  \citep{yang2020improved} and nonparametric case \citep{wu2022integrative}, however extrapolation still requires the function to be uniquely identified from the experimental study.

A strongly related area of work aims to use data from the experimental study to test causal effect estimates from observational studies. One approach aims to falsify causal estimates from observational studies using experimental data \citep{hussain2023falsification,hussain2022falsification}. This work converts a variety of assumptions regarding the validity of the observational study, consistency of the CATE across studies and external validity of the RCT into testable statistical hypothesis, which can then be falsified. Another approach aims to estimate how much unmeasured confounding must be present in an observational study for it to be consistent with the RCT \citep{de2024detecting,de2024hidden}. This is achieved using causal sensitivity models \citep{rosenbaum1983assessing}, which uses a single parameter to control the strength of unmeasured confounding. The goal is then to use the RCT to lower bound this parameter. A significant portion of work in both these areas utilises non-parametric tests of conditional moment restrictions \citep{muandet2020kernel}.

\section{THE HARDNESS OF VALIDATING OBSERVATIONAL STUDIES}\label{sec:hardness_validating}

We now provide some theoretical limits on using experimental data to measure the level of unmeasured confounding in an observational study. We do this using the framework of impossible inference \citep{bertanha2020impossible}, a popular tool in econometrics \citep{canay2013testability}. This field studies when it is possible for a non-trivial hypothesis tests to exist for a problem, where trivial hypothesis tests are those which are unable to distinguish \emph{any} alternative from the null.


Impossible inference has been applied within causal inference to show the hardness of conditional independence testing \citep{shah2020hardness}. In our case, the conditional independence $Y \perp E \mid X,T$ would imply a total lack of unmeasured confounding as:
\begin{align*}
    P_{o}(Y\mid X=\bx,T=t ) &= P_{e}(Y\mid X=\bx,T=t,E=e)\\
    &= P(Y(t)\mid X=\bx,E=e).
\end{align*}
Therefore, the hardness of this testing problem already demonstrates that there are no non-trivial tests for full unconfoundedness in the observational distribution.

However, this still doesn't preclude us from being able to estimate CATE in an unbiased manner from the observational dataset. Or, failing full estimation, bounding the correction function, $\hat{\Delta}(\cdot)$ to return intervals for CATE. Therefore, in this section, we apply the same techniques to focus on what experimental data allows us to test for in term unbiasedness of CATE estimates. This translates to estimating or bounding $\hat{\Delta}(\bx)$ with confidence guarantees, which we do via the following definition:
\begin{definition}\label{def:control}
    We say that that $\hat{\Delta}(\cdot)$ is \textbf{controlled} by functions $\fhigh,\flow: \cX \to \mathbb{R}$ if we have:
 \begin{align}\label{eq:bounding_interval}
    \hat{\Delta}(\bx) \in \squarebrac{ \ubar{f}(\bx),\fhigh(\bx) } \text{ a.s for $ x \sim P_{o}(\bx)$ } 
\end{align}  
Where $\flow(\bx) \leq \fhigh(\bx)$ for all $\bx \in \cX$.
\end{definition}
 If the unmeasured confounding were controlled by $\flow(\bx) = \fhigh(\bx) = 0$ for all $\bx \in \cX$ then we would have that the CATE estimate from the observational study is unbiased to the true CATE. Going further than this, if we knew functions that controlled the confounding, then we could use them alongside the CATE estimate from the observational study to give intervals which contain the true CATE. Therefore, the goal is to understand how we can use the observational study to learn the functions, $\fhigh,\flow$, that control the confounding.

\subsection{Testing Notation and Background}\label{sec:testing_notation_background}
First, we will assume a fixed propensity score function $\pi:\cX \to [0,1]$. Now we define $\cE_{M,\pi}$ to be the set of distributions over $(X,T,Y)$ which are absolutely continuous in $X$ with respect to Lebesgue measure, bounded above in $\ell_{\infty}$ norm by $M$, and whose propensity score is given by $\pi$. As before, the domains of $T, Y, E$ are $\{0,1\}, \mathbb{R}$ and $\{o,e\}$ respectively. For this section, we will use the notation $\mathbb{P}_{P}$ to denote a probability taken with respect to $P \in \cE_{M,\pi}$.

A potentially randomised test $\psi_n$ is a measurable function which takes in a dataset, $\cD$, a random variable $U \sim U\squarebrac{0,1}$ that represents the randomness of the test---a choice of permutations in permutation testing, for example---and outputs a result in $\setbrac{0,1}$ where $1$ corresponds to a rejection. So we write $\psi_n(\cD,U)$ for the result of the test $\psi_n$ with dataset $\cD$ and $U$ as input.  
%


\begin{figure*}[ht]
    \centering
\includegraphics[width=0.9\textwidth]{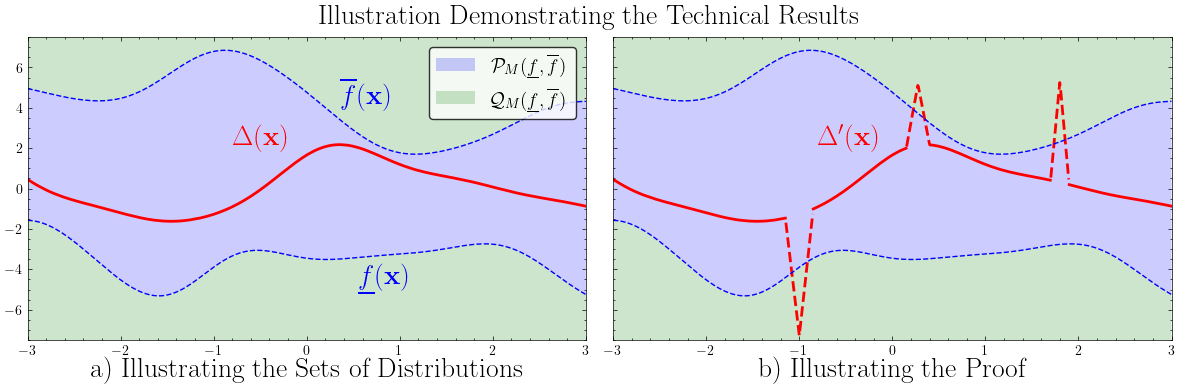}
    \caption{Illustration of both the sets of distributions and proof of the technical result in Section \ref{sec:hardness_validating}. The first figure demonstrates the sets of distributions $\cP_{M,\pi}(\flow,\fhigh),\cQ_{M,\pi}(\flow,\fhigh)$. $\cP_{M,\pi}(\flow,\fhigh)$ is the set of distributions where $\Delta(\cdot)$ is always contained in the blue region, and $\cQ_{M,\pi}(\flow,\fhigh)$ is the set of all other distributions, so those where $\Delta(\cdot)$ leaves the blue region. To prove the hardness of validating observational study estimates, we show that for any $P \in \cP_{M,\pi}(\flow,\fhigh)$ we can find distributions $Q \in \cQ_{M,\pi}(\flow,\fhigh)$ that are arbitrarily close by adding spikes as in Figure b.}
    \label{fig:proof_illustration}
\vspace{-1em}
\end{figure*}

Suppose we observe an experimental dataset $\cD$ sampled i.i.d from a distribution $P_0 \in \cE_{M,\pi}$ and wish to test the null hypothesis $H_0:P_0 \in \cN \subset \cE_{M,\pi}$ against the alternative hypothesis $H_1: P_0 \in \cA \subset \cE_{M,\pi}$. Then we have the following important definitions:
\begin{definition}
    Let $\psi_n$ be a randomised test which takes in a dataset $\cD$ of size $n$. We say $\psi_n$ has \textbf{level} $\alpha$ at size $n$ if we have $\sup_{P \in \cN} \mathbb{P}_{\cD \sim P^n} \circbrac{ \psi_n(\cD,U) =1 } \leq \alpha$. For an alternative distribution $P \in \cA$, we define $\mathbb{P}_{\cD \sim P^n} \circbrac{ \psi_n(\cD,U) =1 }$ as the \textbf{point-wise power} against $P$ at size $n$.
\end{definition}
Ideally, we would want a test to have power against as many alternatives as possible, preferably uniformly so that $\inf_{P \in \cA} \mathbb{P}_{\cD \sim P^n} \circbrac{ \psi_n(\cD,U) =1 } \to 1$. For non-parametric hypothesis, restrictions on the alternative such as smoothness conditions are often required to achieve this \citep{balakrishnan2019hypothesis}.

A particular problem given by sets $(\cN,\cA)$ is known as \textbf{untestable} if for all tests the point-wise power is bounded by the level for any alternative. In this case, there exists no test that can distinguish the null from \emph{any} alternative, therefore the null has to be restricted for there to be an informative test.
\subsection{Setting the Testing Problem for Unmeasured Confounding}\label{sec:testing_problems}

We now apply the above to testing for bias in treatment effect estimation due to unmeasured confounding. Fixing the observational estimate $\hat{\omega}(\cdot)$ and so the correction function $\hat{\Delta}(\cdot)$, we define the following sets of distributions:
\begin{definition}\label{def:distribution_sets}
    Let $\flow,\fhigh: \cX \to \mathbb{R}$. We define:
\begin{align*}
    \cP_{M,\pi}(\flow,\fhigh) &= \setbrac{ P \in \cE_{M,\pi} : {  \text{  $\hat{\Delta}(\cdot)$ controlled by $\flow,\fhigh$ } }} \\
    \cQ_{M,\pi}(\flow,\fhigh) &= \cE_{M,\pi} \setminus \cP_{M,\pi}
\end{align*}  
\end{definition}

\paragraph{Which way round to test?} Now with both sets of distributions , the question is what to take as the null and alternative? We have the following choices:
\begin{align*}
    \text{Test 1:} \begin{cases}
        H_0: P \in \cP_{M,\pi}(\flow,\fhigh) \vspace{0.5em}, &H_1: P \in \cQ_{M,\pi}(\flow,\fhigh),
    \end{cases}  \\
    \text{Test 2:} \begin{cases}
        H_0: P \in \cQ_{M,\pi}(\flow,\fhigh) \vspace{0.5em}, &H_1: P \in \cP_{M,\pi}(\flow,\fhigh),
    \end{cases}
\end{align*}

Rejecting under test $1$ is more standard, and would correspond to \emph{falsifying the hypothesis} that the observational study has a level of confounding controlled by $\flow,\fhigh$. This would mean finding statistical evidence that bias in the observational CATE estimate is not contained in the intervals given by $(\flow,\fhigh)$. 

However, failing to reject under test $1$ \emph{does not provide evidence} that the confounding is controlled by $(\flow,\fhigh)$. We may fail to reject because of other reasons, such as a lack of data or the test having limited power against the true distribution. In an ideal world, we would like to reject the hypothesis that confounding is above a certain level. Fixing this level using some $\flow,\fhigh$, this would correspond to test 2, where the data is used to reject the hypothesis that $\hat{\Delta}$ is not controlled by $\flow,\fhigh$.

The formulation of test 2 is strongly related to bioequivalence testing\footnote{Also referred to as simply equivalence testing.} \citep{wellek2002testing,chow2008design}. In this field, the goal is to find statistical evidence that one medical treatment works almost equivalently to another, where there is some tolerance specified due to working with finite samples. This is used to approve generic drugs, which have the same active ingredient as a branded drug but can only be sold once the branded drug's patent expires. Here the aim is to ensure the generic drug works as well as the branded one, and so consumers can use them interchangeably. 

In our context, we have a similar goal, in that we would like to show that the observational CATE estimate with the adjustment from the RCT is ``good enough" up to some tolerance. We specify this tolerance by functions, $(\flow,\fhigh)$, that control $\Delta$ as in Definition \ref{def:control}.

Following this discussion we provide the following definitions, we refer to tests of type 1 as \textbf{falsification tests} and tests of type 2 as \textbf{equivalence test}.

\subsection{Limits on Testing}\label{sec:limits_testing}

Having laid out the two testing problems in Section \ref{sec:testing_problems}, we now apply the impossible inference framework detailed in Section \ref{sec:testing_notation_background} to these problems. Firstly, we demonstrate that whilst equivalence testing is more aligned with the aim of the field, the problem as set out is untestable:

\begin{theorem}\label{thm:untestable}
Fix any $\flow,\fhigh: \cX \to \mathbb{R}$ and let $\psi_n$ be an equivalence test with null $Q_{M}(\flow,\fhigh)$ and alternative $\cP_{M,\pi}(\flow,\fhigh)$. If the level of this test is, $\alpha$ we have that:
\begin{align}
    \mathbb{P}_P(\psi_n = 1) \leq \alpha,
\end{align}
for any $P \in \cP_{M,\pi}(\flow,\fhigh)$. That is $\psi_n$ does not have power against any alternative. 
\end{theorem}
This shows that \emph{any} equivalence test which has level $\alpha$ against the null will fail to distinguish \emph{any} alternative. This means that there is no hypothesis test that can confirm from data that the difference function $\Delta$ is controlled by any pair of functions $(\flow,\fhigh)$. We visualise the proof of Theorem \ref{thm:untestable} in Figure \ref{fig:proof_illustration}. The idea is that for any distribution $P \in \cP_{M,\pi}(\flow,\fhigh)$, we can construct a series of distributions $\setbrac{Q_i}^\infty_{i=1}: Q_i \in \cQ_{M,\pi}(\flow,\fhigh)$ that tends to $P$ in total variation distance. Therefore, not statistical procedure can distinguish the two.

This result has implications for falsification tests:
\begin{corollary}\label{cor:untestable}
For fixed $\flow,\fhigh: \cX \to \mathbb{R}$, any falsification test $\psi_n$ with null $\cP_{M,\pi}(\flow,\fhigh)$ and alternative $\cQ_{M,\pi}(\flow,\fhigh)$ has:
\begin{align}
    \inf_{Q \in \cQ_{M,\pi}(\flow,\fhigh)} \mathbb{P}_Q(\psi_n = 1) \leq \alpha,
\end{align}
where $\alpha$ is the level of $\psi_n$.
\end{corollary}
This means that for any $n$, any falsification test will always fail to distinguish some set of alternatives from the null with power distinctly above the level. However, the next result shows there are alternatives which can be distinguished from the null in falsification tests:
\begin{proposition}\label{prop:falsifiable}
There exists a distribution $Q \in \cQ_{M,\pi}(\flow,\fhigh)$ such that $\mathrm{TV}(Q,\mathrm{co}(\cP_{M,\pi}(\flow,\fhigh)))\geq \beta$ for some $\beta>0$ where $\mathrm{co}(\cP_{M,\pi}(\flow,\fhigh)$  is the convex hull of  $\cP_{M,\pi}(\flow,\fhigh)$. Following \citet{bertanha2020impossible}, this guarantees that there is a test with $\beta+\alpha$ against $Q$ where $\alpha$ is the level of the test.   
\end{proposition}

This demonstrates that, unlike validation, falsification is possible relative to certain alternatives.  
\subsection{Implications for Sensitivity Models}

Finally, we apply the results in Section \ref{sec:limits_testing} to approaches that make use of causal sensitivity models to measure the degree of unmeasured confounding. First, defining a generalised sensitivity model as follows:
\begin{definition}
    A \textbf{sensitivity model} is a parameterised set of pairs of functions:
    \begin{align}
\setbrac{(\flow_{\gamma},\fhigh_{\gamma}):\flow_{\gamma},\fhigh_{\gamma}: \cX \to \mathbb{R}, \gamma \in \squarebrac{\Gamma_0,\Gamma_1}},
    \end{align} where for fixed $\bx$, $\flow_{\gamma}(\bx),\fhigh_{\gamma}(\bx)$ are continuously decreasing/increasing respectively in $\gamma$. Moreover, that $\flow_{\Gamma_0}(\bx),\fhigh_{\Gamma_0}(\bx)=0$ and  $\flow_{\Gamma_1}(\bx),\fhigh_{\Gamma_1}(\bx)$ are $-M$ and $M$ respectively. For a distribution in $P \in \cE_M$ we define:
    \begin{align}
        \Gamma(P) = \inf \setbrac{\gamma: (\flow_{\gamma},\fhigh_{\gamma}) \text{ controls } \Delta(\cdot)}
    \end{align}
\end{definition}
Viewed this way, a sensitivity model is a way of constructing intervals around the confounded CATE, $\omega_o(\bx)$, that contain the true CATE. Moreover, previous work in this area can be seen as aiming to use the experimental dataset to perform inference on $\Gamma(P)$. Specifically, \citet{de2024detecting,de2024hidden} both look to use the experimental data to construct probabilistic lower bounds on $\Gamma(P)$. We now apply results Section \ref{sec:limits_testing} to constructing confidence intervals for $\Gamma(P)$, showing non-trivial upper bounds are not possible:
\begin{theorem}\label{thm:sensativity}
    Fix a sensitivity model and let $\cD$ be a dataset sampled from $P^{(n)}$ where $P \in \cE_{M,\pi}$.  Let $\squarebrac{\ubar{C}\circbrac{\cD},\bar{C}\circbrac{\cD}}$ be a confidence interval for $\Gamma(P)$ in that it satisfies the following coverage requirement:
\begin{align}
    \inf_{P\in \cE_{0,M}} \PP_{\cD \sim P^{(n)}} \circbrac{\Gamma(P) \in C\circbrac{\cD_n}} \geq 1-\alpha
\end{align}
Then $\bar{C}\circbrac{\cD} = \Gamma_1$ with probability $1-\alpha$. That is, there are no non-trivial upper bounds on $\Gamma(P)$.
\end{theorem} 
This demonstrates that in contrast to the lower bound case, non-trivial probabilistic upper bounds on $\Gamma(P)$ are not possible without further assumptions on the set of distributions, $\cE_M$. This creates difficulties in using sensitivity models to benchmark unmeasured confounding, as a lower bound represents the smallest amount of confounding that can explain the data. Therefore, it \emph{does not} allow us to say with confidence that a treatment effect is contained in some interval.

\section{PSEUDO-OUTCOME GAUSSIAN PROCESSES AND UNIFORM ERROR BOUNDS}

The results presented in Section \ref{sec:hardness_validating} show that without further assumptions, we cannot produce intervals which contain the true CATE with high probability. The proof relied on constructing arbitrary peaks in the correction function, which was possible due to a lack of smoothness. As an example of an assumption that can permit this type of inference, we leverage Gaussian process (GPs) which come with inherent smoothness constraints. We then adapt uniform error bounds for Gaussian processes \citep{fiedler2021practical,lederer2019uniform} in order to get functions which control $\Delta(\cdot)$.

Before doing this, we develop a Gaussian process approach to learning CATE from pseudo-outcomes. To this best of our knowledge, this is first example of such a method. This is important as it allows us to learn the difference function directly from an estimate of $\omega_o(\bx)$ in the observational study. Alternative causal Gaussian process approaches, such as \citet{alaa2017bayesian}, would require access to an estimate of $\mathbb{E}\squarebrac{Y\mid X=\bx, T=t,E=o}$, to learn a separate correction for each $t$. By using the pseudo-outcome approach, we sidestep this issue and so allow users full flexibility on how to model $\omega_o(\bx)$.


\subsection{Pseudo Outcome Regression with Gaussian Processes}\label{sec:pseudo_GP}

We now turn to designing a GP based pseudo-outcome approach. pseudo-outcomes are designed so that the minimiser of the pseudo-outcome mean squared error is the same as the minimiser of the mean squared error under the true unobserved CATE. If the propensity score is correctly specified, the pseudo mean square error will converge optimally to the true CATE mean square error \citep{kennedy2023towards}. However, GP based methods are fit via maximum likelihood, with closed form solutions to the posterior requiring Gaussian errors. This creates a problem for applying them directly to pseudo-outcomes, which have distinctly non Gaussian errors. Specifically, they may be written as:
\begin{align}
    \Tilde{Y} = \tau(X) + \epsilon
\end{align}
Where $\EE[\epsilon \mid X] = 0$ but $\EE[\epsilon \mid X,T] \neq 0$. As we show in Appendix \ref{ap:guassian_process}, this breaks Gaussianity assumptions even in cases where the errors on $Y$ are Gaussian. However, the next proposition suggests a simple correction term which allows us to recover well-behaved errors:
\begin{figure*}[ht]
    \centering
\includegraphics[width=\textwidth]{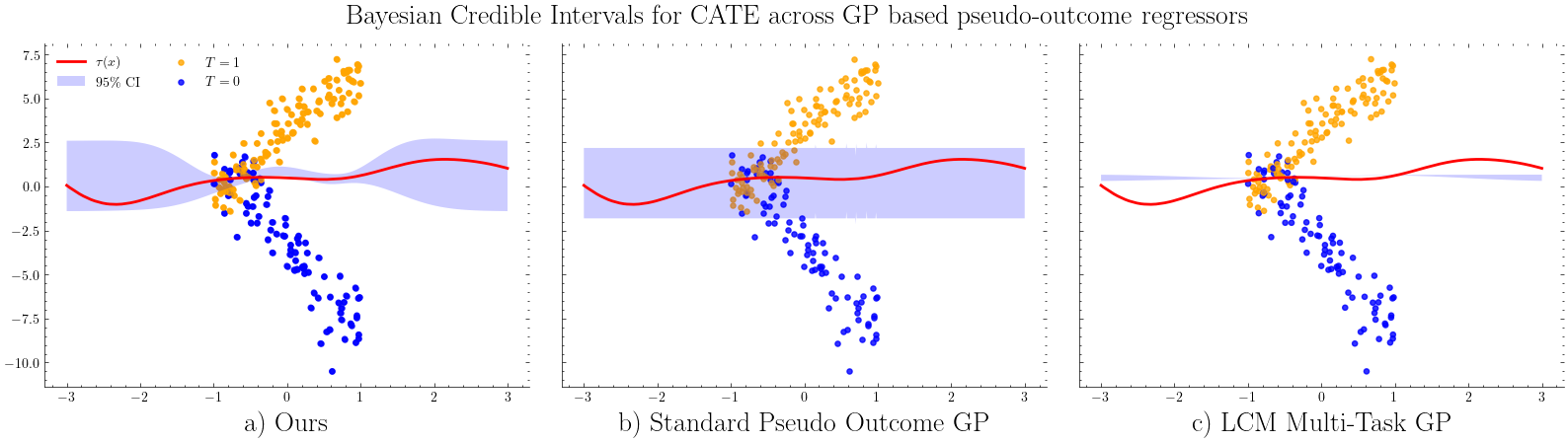}
    \caption{A particularly pathological example of the behaviour we observe for each method in our simulated experiment of Section \ref{sec:simulated_experiment}. For the standard GP, hyperparameter optimisation leads  to uninformative predictions as it cannot account for close $\bx$ values with seemingly no correlation. For the trained LCM, we get strong predictive performance but poor uncertainty quantification, especially out of distribution. Our approach gets the best of both scenarios, with strong predictive performance and calibrated uncertainty out of distribution. }
    \label{fig:simulated_plot}
\vspace{-1em}
\end{figure*}
\begin{proposition}\label{prop:error_dist}
There exists a function $\phi: \cX \to \mathbb{R}$ such that the IPW pseudo-outcome can be written as:
\begin{align*}
    \Tilde{Y} = \tau(X) + \circbrac{\frac{T-\pi(X)}{\pi(X)\circbrac{1-\pi(X)}}} \phi(X) + \Tilde{\epsilon}
\end{align*}
Where $\EE[\Tilde{\epsilon} \mid X,T] = 0$ and $\Tilde{\epsilon}\mid T,X$ is Gaussian if the original errors on $Y$ are.
\end{proposition}

Using the decomposition provided by this proposition we may model this using independent Gaussian Processes for $\hat{\Delta}$ and $\phi$, so $\hat{\Delta} \sim \mathrm{GP}(0, k_{\theta}),\phi \sim \mathrm{GP}(0, l_{\eta})$. This is equivalent to using a vector valued GP \citep{alvarez2012kernels} for multitask regression, where we take the three tasks to be predicting the pseudo-outcome when $T=0$, predicting the pseudo-outcome when $T=1$, and the finally predicting the CATE. This corresponds to using the following LCM multitask kernel \citep{alvarez2012kernels}:
\begin{align}
    \bK &= \ba\ba^{\top} k_{\theta} + \bb\bb^{\top} l_{\eta}\label{eq:matrix_parameters}\\
        \ba &= \begin{bmatrix}
1 & 1 & 1\\
\end{bmatrix} \\
\bb &= \begin{bmatrix}
\frac{-1}{1-\pi(X)} & \frac{1}{\pi(X)} & 0\\
\end{bmatrix}
\end{align}

From here, we can compute the posterior for $\hat{\Delta}$ given the pseudo-outcomes whilst marginalising out the effect of $\phi$. That can all be done in closed form, and gives a posterior of the form $\hat{\Delta}(\bx) \sim GP(\Tilde{\Delta}_{\cD_e}(\cdot),k_{\cD_e}(\cdot,\cdot))$ with expressions for $\Tilde{\Delta}_{\cD_e}(\cdot),k_{\cD_e}(\cdot,\cdot)$ in Appendix \ref{ap:guassian_process}.
\subsection{Uniform Error Bounds}

We now turn to the main purpose of using Gaussian Processes in that we provide a set of assumptions under which our method we can learn functions that control $\Delta(\cdot)$ from experimental data. We do this by adapting the uniform error bounds for GPs from \citet{lederer2019uniform} to our specific model. First, providing the assumption which makes inference possible:

\begin{assumption}\label{assump:GP}
The unknown $\hat{\Delta}$ and $\phi$ are samples from Gaussian processes with kernel $k$ and $l$ respectively, i.e $\hat{\Delta}(\cdot) \sim \mathrm{GP}(0,k)$ and $\phi \sim \mathrm{GP}(0,l)$. Further we assume $\cX_o$ is compact, the errors have distribution $\cN(0,\sigma_t^2)$ given $T=t$, $k$ has Lipschitz constant $L_k$, and $\hat{\Delta}$ has a Lipschitz constant $L_{\hat{\Delta}}$. 
\end{assumption}

This implies the following bounds on the $\hat{\Delta}(\cdot)$:
\begin{theorem}\label{thm:uniform_bounds}
Let the posterior for $\hat{\Delta}(\cdot)$ from the GP model defined in Section \ref{sec:pseudo_GP} be given pointwise by $\cN(\Tilde{\Delta}(\bx),\sigma^{2}(\bx))$ where $\Tilde{\Delta},\sigma: \cX_o \to \mathbb{R}$ and let $\hat{\tau}(\bx) = \hat{\omega}(\bx)+ \Tilde{\Delta}(\bx)$. Then, under assumption \ref{assump:GP}, for fixed $\delta \in (0,1),\tau\in \mathbb{R}^{+}$ we have:
\begin{align*}
    P(\lvert\hat{\tau}(\bx) - {\tau}(\bx) \rvert& \leq B(\bx) \hspace{0.5em} \forall \bx \in \cX_o) > 1- \delta \\
    B(\bx) &= \sqrt{2\log \circbrac{\frac{M\circbrac{\tau,\cX_o}}{\delta}}}\sigma(\bx) \\
    &+ \gamma(\tau,\bX_e,L_k,L_{\Delta})
\end{align*}
Where $M\circbrac{\tau,\cX_o}$ is the $\tau$ covering number of $\cX_o$, defined as the minimum number of spherical balls of radius $\tau$ needed to cover $\cX_o$, and $\gamma(\tau,\bX_e,L_k,L_{\Delta})$ is defined in Appendix \ref{ap:guassian_process}.
\end{theorem}
There is not a uniformly optimal value of $\tau$ due to dependence on other constants. However, if our covariate space is a hypercube of length $r$ with dimension $d$, we have that $M\circbrac{\tau,\cX_o}\leq (1+\frac{r}{\tau})^d$ and as we show in Appendix \ref{ap:guassian_process}, we have that $\gamma(\tau,\bX_e,L_k,L_{\Delta}) = {o}(\tau^{\frac{1}{2}})$. This ensures we can always recover log dependency in $\delta$ and $\tau$. Further, the kernel choice gives us knowledge of $L_k$ and allows us to probabilistically bound $L_f$ as in \citet{lederer2019uniform} and shown in Appendix \ref{ap:guassian_process}.
\section{Experiments}
For the experiments, we demonstrate the improvements in predictive performance and calibrated uncertainty provided by our pseudo-outcome GP approach when compared against other causal GP approaches. We compare against fitting a naive standard GP, and a GP with a multitask LCM kernel to pseudo-outcomes. This second approach which can be viewed as a scaled version of the causal multitask Gaussian process of \citep{alaa2017bayesian},which represents the state of the art in GP's for CATE estimation\footnote{For more on the comparison between the LCM GP and Causal Multitask GP's see Appendix \ref{ap:LCM_comparison}}. For all models we tune the free hyperparameters using gradient descent on the marginal log likelihood, details in Appendix \ref{ap:model_exp_details}. For results on the coverage of uniform error bounds for our model specifically, see \ref{ap:uniform_error_bounds}.

\subsection{Simulated Experiment}\label{sec:simulated_experiment}

Firstly, we use an adaptation of the simulated provided in \citet{kallus2018removing}. We let the experimental and observational covariate distribution be $\cU([-1,1]^d)$ and $\cU([-3,3]^d)$ respectively, where $d$ is the covariate dimension, and $T \sim \mathrm{Ber}(\frac{1}{2})$. The observational outcomes are simulated using a quadratic in the first component of $X$ and $T$ with normal noise. For the experimental outcomes, we simulate use the same polynomial but add a sample from a GP. We do this in order to test our methodology in setting where assumptions are satisfied. Full details are given in Appendix \ref{ap:experiment_details}. 

Finally, as we focus on assessing the GP portion of the model, we use the true $\omega_o(\bx)$ for this experiment. This represents a case where $n_o$ is so large, we approach fitting a perfect model. We later relax this.
\begin{table}
    \centering
     \resizebox{\linewidth}{!}{\begin{tabular}{|c|c|c|c|}
    \hline
    \textbf{Model} &\textbf{MSE}  &\textbf{Coverage} & \textbf{Interval Width}\\
     \hline
        Ours & $\mathbf{1.77 \pm 0.01}$ & ${0.785 \pm 0.04}$  & $\mathbf{3.31 \pm 0.02}$ \\
         \hline
        Naive GP & ${2.05 \pm 0.02}$  & $\mathbf{0.796 \pm 0.04}$  & $3.65 \pm 0.03$ \\
         \hline
        LCM & ${1.91 \pm 0.01}$ & ${0.303 \pm 0.11}$ & ${1.09 \pm 0.06}$ \\
         \hline
    \end{tabular}}
    \caption{Results for the simulated experiment in Section \ref{sec:simulated_experiment} with $d=10$ and $n_e=1000$, averaged over 200 runs. Our approach leads to both the best predictive performance and well calibrated uncertainty, achieving a similar coverage to the standard GP with smaller predictive intervals. }
    \label{tab:sim_results_main}
\vspace{-1em}
\end{table}
\subsubsection{Results}

We present results for this experiment with $d=10$ and $n_e=1000$ in Table \ref{tab:sim_results_main}, alongside an illustrative figure for $d=1$ and $n_e=200$ in Figure \ref{fig:simulated_plot}. In Table \ref{tab:sim_results_main} we compare the mean squared error to the true CATE, the coverage of $95\%$ Bayesian credible intervals, and the width of these intervals. We use Bayesian credible intervals to form a fair comparison for uncertainty quantification, as uniform error bounds are only available for our model. Additional results including varying dimension, varying sample size, and extrapolation beyond the experimental sample in Appendix \ref{ap:simulated_additional_results}.  

Across all settings, our results show the following trends: i) The naive GP shows poor predictive performance and is totally uninformative in some settings. This is because it is unable to correctly optimise hyperparameters to capture the highly variable noise distribution in the pseudo-outcomes, and so it reverts to the prior ii) The trainable LCM multitask kernel has good predictive performance but poorly calibrated uncertainty, This is because the hyperparameter optimisation either leads to overly confident or overly wide credible intervals, depending on the dimension and the number of samples. Further, this optimisation leads it to overfit training data and extrapolate poorly. iii) Our approach is able to incorporate both strong predictive performance and calibrated uncertainty, with intervals that have the coverage guarantees of other methods that produce much wider intervals. 
\begin{table}
    \centering
     \resizebox{\linewidth}{!}{\begin{tabular}{|c|c|c|c|}
    \hline
    \textbf{Model} &\textbf{MSE}  &\textbf{Coverage} & \textbf{Interval Width}\\
     \hline
        Ours & $\mathbf{1.10 \pm 0.04}$ & $\mathbf{0.831 \pm 0.008}$  & $\mathbf{2.66 \pm 0.02}$ \\
         \hline
        Naive GP & ${2.19 \pm 0.13}$  & ${0.752 \pm 0.010}$  & $3.38 \pm 0.03$ \\
         \hline
        LCM & ${1.39 \pm 0.05}$ & ${0.828 \pm 0.010}$ & ${3.00 \pm 0.05}$ \\
         \hline
    \end{tabular}}
    \caption{Results for the IHDP setting described in Section \ref{sec:semi_synthetic} with $n_{e} = 400$ averaged over 100 runs. We again find that our method has the best predictive performance and most informative coverage intervals, in the sense that they contain the true CATE with high probability whilst also being significantly smaller. }
    \label{tab:IDHP_results_main}
\vspace{-1em}
\end{table}
\subsection{Semi-Synthetic Experiments}\label{sec:semi_synthetic}
To assess our method in a more realistic setting, we use the Infant Health and Development Program (IHDP) dataset \citep{louizos2017causal}, similarly to \citet{hussain2022falsification,hussain2023falsification}. The IHDP dataset comes from a randomised controlled trial, and it contains $n=985$ samples and a $28$ dimensional covariate distribution with $7$ continuous covariates. The dataset comes with a treatment allocation, but outcomes need to be simulated. 

We form our data from the IHDP dataset as follows: for the observational sample, we uniformly sample the covariates and treatment with replacement until reaching the desired sample size. For the experimental study, we do a weighted sampling to ensure a covariate shift and then randomly sample the treatment from $T \sim \mathrm{Ber}(p)$. For the outcomes in the observational dataset, we simulate from a sparse linear model in $X$ and $T$. We do this as we want to emulate a scenario where we have relatively low error in estimating $\omega_o(\bx)$ due to a large $n_o$, but we do not want to repeat the small dataset multiple times. Finally, we again simulate the difference between observational and experimental distributions with a GP as in Section \ref{sec:simulated_experiment}. Full details available in Appendix \ref{ap:IHDP_details}.

\subsubsection{Results}

We present the results for the experiment with $n_e=400$ in Table \ref{tab:IDHP_results_main}, with other results for varying sample size, treatment proportion and out of distribution generalisation in Appendix \ref{ap:idhp_additional_results}. We can see that our proposed methodology outperforms both alternatives in terms of predictive performance and calibrated uncertainty, having the joint best coverage with the smallest intervals. The advantage becomes even more clear when extrapolating beyond the experimental study to the observational study, as shown in Table \ref{ap:idhp_additional_results} in Appendix \ref{ap:idhp_additional_results}. In this case, the trained GP predicts overly broad intervals due to over-fitting the hyperparameters to the experimental data.

\subsection{Additional Results}
Finally, we highlight some additional results available in the Appendices. In Appendix \ref{ap:simulated_additional_results}, we present our simulated experiment over different dimensionality and sample sizes, in Appendix \ref{ap:idhp_additional_results} we present the IHDP study for different treatment proportions, and for a varying quality of fit of the observational model and in Appendix \ref{ap:robustness_results} we present results for both scenarios, varying the difference function so that it is not a GP. Finally, in Appendix \ref{ap:robustness_results}, we provide results on our uniform error bounds for our proposed GP model. 

\section{CONCLUSION}
It is well known in causal inference that there are no observational tests for unmeasured confounding. In this work, we showed that even with experimental data, there are fundamental limits to testing for unmeasured confounding. We showed that in order to validate observational studies, one needs to make assumptions on the smoothness of the correction function. Following this, we developed a Gaussian Process approach to learning from pseudo-outcomes, and assumptions arising from this model which produce intervals that contain the CATE with high probability.

\section*{Acknowledgements}
JF gratefully acknowledges funding from the EPSRC.
 
\bibliography{references}
\bibliographystyle{abbrvnat}

\section*{Checklist}

 \begin{enumerate}

 \item For all models and algorithms presented, check if you include:
 \begin{enumerate}
   \item A clear description of the mathematical setting, assumptions, algorithm, and/or model. [Yes]
   \item An analysis of the properties and complexity (time, space, sample size) of any algorithm. [Yes]
   \item (Optional) Anonymized source code, with specification of all dependencies, including external libraries. [No]
 \end{enumerate}

 \item For any theoretical claim, check if you include:
 \begin{enumerate}
   \item Statements of the full set of assumptions of all theoretical results. [Yes]
   \item Complete proofs of all theoretical results. [Yes]
   \item Clear explanations of any assumptions. [Yes]     
 \end{enumerate}

 \item For all figures and tables that present empirical results, check if you include:
 \begin{enumerate}
   \item The code, data, and instructions needed to reproduce the main experimental results (either in the supplemental material or as a URL). [Yes]
   \item All the training details (e.g., data splits, hyperparameters, how they were chosen). [Yes]
         \item A clear definition of the specific measure or statistics and error bars (e.g., with respect to the random seed after running experiments multiple times). [Yes]
         \item A description of the computing infrastructure used. (e.g., type of GPUs, internal cluster, or cloud provider). [Yes]
 \end{enumerate}

 \item If you are using existing assets (e.g., code, data, models) or curating/releasing new assets, check if you include:
 \begin{enumerate}
   \item Citations of the creator If your work uses existing assets. [Not Applicable]
   \item The license information of the assets, if applicable. [Not Applicable]
   \item New assets either in the supplemental material or as a URL, if applicable. [Not Applicable]
   \item Information about consent from data providers/curators. [Not Applicable]
   \item Discussion of sensible content if applicable, e.g., personally identifiable information or offensive content. [Not Applicable]
 \end{enumerate}

 \item If you used crowdsourcing or conducted research with human subjects, check if you include:
 \begin{enumerate}
   \item The full text of instructions given to participants and screenshots. [Not Applicable]
   \item Descriptions of potential participant risks, with links to Institutional Review Board (IRB) approvals if applicable. [Not Applicable]
   \item The estimated hourly wage paid to participants and the total amount spent on participant compensation. [Not Applicable]
 \end{enumerate}

 \end{enumerate}

\appendix

\onecolumn

\section{Causal Assumptions}

\subsection{Constant CATE}\label{ap:constant_CATE}

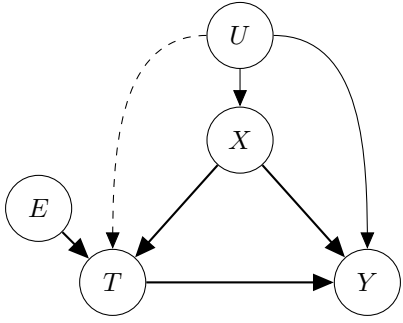
\begin{figure}
    \centering
\begin{tikzpicture}
    \node (T) [state, standard_edge_color, fill=standard_node_color] {$T$};
    \node (mid) [state, standard_edge_color, fill=standard_node_color, right=\horizontalsep of T,draw=none] {};
    \node (X) [state, standard_edge_color, fill=standard_node_color, above=\verticalsep of mid] {$X$};
    \node (Y) [state, standard_edge_color, fill=standard_node_color, right= \horizontalsep of mid] {$Y$};
    \node (U) [state, standard_edge_color, fill=standard_node_color, above= 0.5*\verticalsep of X] {$U$};
    \node (E) [state, standard_edge_color, fill=standard_node_color, above left= 0.5*\verticalsep of T] {$E$};
    \draw[->, thick, standard_edge_color] (T) edge (Y);
    \draw[->, thick, standard_edge_color] (E) edge (T);
    \draw[->, thick, standard_edge_color] (X) edge (Y);
    \draw[->, thick, standard_edge_color] (X) edge (T);
    \draw[->, dashed, standard_edge_color,in =90,out = 180] (U) edge (T);
    \draw[->, standard_edge_color,in= 90, out = 0] (U) edge (Y);
    \draw[->, standard_edge_color] (U) edge (X);
    \end{tikzpicture}
    \caption{Causal Structure for generating the experimental and observational datasets with environment node drawn in.}
    \label{fig:generating_DAG_env_node}
\end{figure}

Drawing the graph with an environment node in Figure \ref{fig:generating_DAG_env_node}, we can read off the graph using the SWIG framework \citep{richardson2013single} for conditional indecencies of potential outcomes from graphical models that $Y(t) \perp E \mid X$. This implies:
\begin{align}
    \EE\squarebrac{Y(t)\mid X,E=e} =\EE\squarebrac{Y(t)\mid X,E=o}
\end{align}
And so constant CATE.

\section{Hardness of Testing}\label{ap:hard_test}

\begin{customthm}{\ref{thm:untestable}}
Fix any $\flow,\fhigh: \cX \to \mathbb{R}$ and let $\psi_n$ be an equivalence test with null $\cQ_{M,\pi}(\flow,\fhigh)$ and alternative $\cP_{M,\pi}(\flow,\fhigh)$. If the level of this test is $\alpha$ we have that:
\begin{align}
    \mathbb{P}_P(\psi_n = 1) \leq \alpha,
\end{align}
for any $P \in \cP_{M,\pi}(\flow,\fhigh)$. That is $\psi_n$ does not have power against any alternative. 
\end{customthm}

\begin{proof}
        Following \citet{romano2004non}, we need to show that the null is dense in the alternative in total variation distance. This corresponds to showing that for $P \in \cP_{M,\pi}(\flow,\fhigh)$ we can find a sequence of distributions $\setbrac{Q_n}^{\infty}_{n=1} \in \cQ_{M,\pi}(\flow,\fhigh)$ such that $d_{\mathrm{TV}}(P,Q_n) \to 0$. 
        
Firstly, note that as $Y$ is bounded by $M$ we must have that $\norm{\hat{\Delta}(\bx) +\hat{\omega}_{o}(\bx)} \leq 2M$ which implies $-\circbrac{\hat{\omega}_o(\bx) +2M} \leq   \hat{\Delta}(\bx)\leq 2M -\hat{\omega}_o(\bx)$. So in order for the bounds to be non vacuous we need $\fhigh(\bx) < 2M - \hat{\Delta}(\bx)\leq\hat{\omega}_o(\bx)$ or $\flow(\bx) > -\circbrac{\hat{\omega}_o(\bx) +2M}$ on some set of positive measure. Assume wlog that we have $\fhigh(\bx) < 2M -\hat{\omega}_o(\bx)$ and then let $\cA_n = \subset \cX$ be a measurable set such that $P(\cA_n) = \epsilon_n$ where $ 0<\epsilon_n\leq\frac{1}{n^2}$ such that this holds. We can find such a set as $P$ is absolutely continuous in $\bx$ with respect to the Lebesgue measure.

    
    
    Now define $R_n$ as a distribution over $(X,T,Y)$ where $X$ is uniform on $\cA_n$, conditional distribution of $T$ given $X$ coming from $\pi$ and the distribution over $Y$ is such that the true CATE is $2M$. Now we let:
\begin{align}
    Q_n \coloneqq \frac{n-1}{n}P + \frac{1}{n}R_n
\end{align}
Now consider the expectation of $\Tilde{Y}$ under this distribution, given $X \in \cA_n$:
\begin{align}
    \EE_{Q_n}[\Tilde{Y} \mid X \in \cA_n] &= \EE_{P}[ \Tilde{Y} \mid X \in \cA_n]\PP\circbrac{ (\Tilde{Y},X)\sim P \mid X \in \cA_n } \\
    &+ \EE_{R_n}[\Tilde{Y} \mid X \in \cA_n]\PP\circbrac{ (\Tilde{Y},X)\sim R_n \mid X \in \cA_n }
\end{align}
We have the following:
\begin{align}
    \PP\circbrac{ (\Tilde{Y},X)\sim P \mid X \in \cA_n } &= \frac{ \PP\circbrac{  X \in \cA_n \mid (\Tilde{Y},X)\sim P } \PP\circbrac{ (\Tilde{Y},X)\sim P } }{\PP\circbrac{  X \in \cA_n}} \\
    &= \frac{\epsilon_n \frac{n-1}{n} }{\epsilon_n \frac{n-1}{n} + \frac{1}{n}}\\
    &=\frac{\epsilon_n (n-1) }{\epsilon_n (n-1) + 1}
\end{align}
And:
\begin{align}
    \PP\circbrac{ (\Tilde{Y},X)\sim Q_n \mid X \in \cA_n } &= 1 - \PP\circbrac{ (\Tilde{Y},X)\sim P \mid X \in \cA_n } \\
    &=\frac{1 }{\epsilon_n (n-1) + 1}
\end{align}
So that:
\begin{align}
    \EE_{Q_n}[\Tilde{Y} \mid X \in \cA_n] &=  \frac{\epsilon_n (n-1) \EE_{P}[Z \mid X \in \cA_n] }{\epsilon_n (n-1) + 1}  + \frac{\EE_{Q_n}[Z \mid X \in \cA_n]}{\epsilon_n (n-1) + 1} \\
    &=\frac{\epsilon_n (n-1) \EE_{P}[\Tilde{Y} \mid X \in \cA_n] }{\epsilon_n (n-1) + 1}  + \frac{2M}{\epsilon_n (n-1) + 1}
\end{align}
Now as $\epsilon_n (n-1) \to 0$ as $n \to \infty $ we have $\EE_{P_n}[\Tilde{Y} \mid X \in \cA_n] \to 2M$. As we have $\hat{\Delta}(\bx) = \EE_{P}[\Tilde{Y} \mid \bx] -\hat{\omega}_o(\bx)$, we have that $\hat{\Delta}(\bx) \to 2M -\hat{\omega}_{o} (\bx) $ for $\bx \in \cA_n$ . Therefore, by taking the cutoff sequence $\setbrac{Q_n}^{\infty}_{n\geq k}$ for some $k$ we have a sequence of distributions such that  $\hat{\Delta}(\bx)$ expectation is greater than $\flow(\bx)$ on a set of positive measure and that $d_{\mathrm{TV}}(Q,P_n) \to 0$. 
\end{proof}

\begin{customcor}{\ref{cor:untestable}}
For fixed $\flow,\fhigh: \cX \to \mathbb{R}$, any falsification test $\psi_n$ with null $\cP_{M,\pi}(\flow,\fhigh)$ and alternative $\cQ_{M,\pi}(\flow,\fhigh)$ has:
\begin{align}
    \inf_{Q \in \cQ_{M,\pi}(\flow,\fhigh)} \mathbb{P}_Q(\psi_n = 1) \leq \alpha,
\end{align}
where $\alpha$ is the level of $\psi_n$.
\end{customcor}
\begin{proof}
    This follows as a direct result of the fact that the alternative is dense in the null, as shown in the previous proposition
\end{proof}

\begin{customprop}{\ref{prop:falsifiable}}
 There exists a distribution $Q \in \cQ_{M,\pi}(\flow,\fhigh)$ such that $\mathrm{TV}(Q,\mathrm{co}(\cP_{M,\pi}(\flow,\fhigh)))\geq \beta$ for some $\beta>0$ where $\mathrm{co}(\cP_{M,\pi}(\flow,\fhigh)$  is the convex hull of  $\cP_{M,\pi}(\flow,\fhigh)$. Following \citet{bertanha2020impossible}, this guarantees that there is a test with $\beta+\alpha$ against $Q$ where $\alpha$ is the level of the test.   
\end{customprop} 
\begin{proof}
As in the proof of Theorem \ref{thm:untestable} assume wlog that we have $\fhigh(\bx) < 2M -\hat{\omega}_o(\bx)$ for some measurable set $\cA$. Now define a distribution $Q \in \cQ_{M,\pi}(\flow,\fhigh)$ as follows:
\begin{itemize}
    \item $X$ is uniform on $\cA$.
    \item $T|X \sim \pi(X)$.
    \item $Y|T = M(2T-1)$.
\end{itemize}
Now let $\Tilde{A} = \mathrm{supp}(Q)$ be the support of $Q$. We have that $\EE_P[\Tilde{Y}| (X,T,Y)\in \Tilde{\cA} ] = 2M$ for any distribution $P$. Now for $P\in \cQ_{M,\pi}(\flow,\fhigh)$ we have that $\EE_P[\Tilde{Y}| (X)\in \cA ] \leq \fhigh(\bx) < 2M -\hat{\omega}_o(\bx)$. Now we have:
\begin{align}
    \EE_P[\Tilde{Y}| X\in \cA ] &= \EE_P[\Tilde{Y}| (X,T,Y)\in \Tilde{\cA} ] \PP_P((X,T,Y)\in \Tilde{\cA} \mid X \in \cA)\\ 
    &+ \EE_P[\Tilde{Y}| (X,T,Y)\in \cA \setminus \Tilde{\cA} ] \circbrac{1-\PP_P((X,T,Y) \in \Tilde{\cA} \mid X \in \cA)} \\
    &\geq 2M\circbrac{2\PP_P((X,T,Y) \in \Tilde{\cA} \mid X \in \cA) - 1}  
\end{align}
As $\EE_P[\Tilde{Y}| (X,T,Y)\in \cA \setminus \Tilde{\cA} ] \geq -2M$. Putting this together implies:
\begin{align}
    &2M -\hat{\omega}_o(\bx) \geq 2M\circbrac{2\PP_P((X,T,Y) \in \Tilde{\cA} \mid X \in \cA) - 1}  \\
    &\implies 
    \PP_P((X,T,Y) \in \Tilde{\cA} \mid X \in \cA) \leq 1 - \frac{\hat{w}_o(\bx)}{4M}\\
    &\implies 
    \PP_P((X,T,Y) \in \Tilde{\cA}) \leq 1 - \frac{\hat{w}_o(\bx)}{4M}
\end{align}
Which completes the proof.
\end{proof}

\begin{customthm}{\ref{thm:sensativity}}
    Fix a sensitivity model and let $\cD$ be a dataset sampled from $P^{(n)}$ where $P \in \cE_{M,\pi}$.  Let $\squarebrac{\ubar{C}\circbrac{\cD},\bar{C}\circbrac{\cD}}$ be a confidence interval for $\Gamma(P)$ in that it satisfies the following coverage requirement:
\begin{align}
    \inf_{P\in \cE_{0,M}} \PP_{\cD \sim P^{(n)}} \circbrac{\Gamma(P) \in C\circbrac{\cD_n}} \geq 1-\alpha
\end{align}
Then $\bar{C}\circbrac{\cD} = \Gamma_1$ with probability $1-\alpha$. That is, there are no non trivial upper bounds on $\Gamma(P)$.
\end{customthm} 
\begin{proof}
    This follows from the fact that for any $\gamma \in \squarebrac{\Gamma_0,\Gamma_1}$ we have that $\cQ_{M,\pi}(\flow_{\gamma},\fhigh_{\gamma})$ is dense in  $\cP_{M,\pi}(\flow_{\gamma},\fhigh_{\gamma})$. Therefore, for any distribution $P \in \cE_{0,M}$ is arbitrarily close in total variation to a distribution whose true sensativity parameter is arbitrarily high. Therefore, if we are to satisfy the coverage requirment uniformly over all distributions we must have $\bar{C}\circbrac{\cD} = \Gamma_1$ with probability $1-\alpha$.
\end{proof}

\section{Gaussian Process}\label{ap:guassian_process}
\begin{customprop}{\ref{prop:error_dist}}
There exists a function $\phi: \cX \to \mathbb{R}$ such that the IPW pseudo-outcome can be written as:
\begin{align*}
    \Tilde{Y} = \tau(X) + \circbrac{\frac{T-\pi(X)}{\pi(X)\circbrac{1-\pi(X)}}} \phi(X) + \Tilde{\epsilon}
\end{align*}
Where $\EE[\Tilde{\epsilon} \mid X,T] = 0$ and $\Tilde{\epsilon}\mid T,X$ is Gaussian if the original errors on $Y$ are.
\end{customprop}
\begin{proof}
   Suppose we have:
\begin{align}
    Y = \mu(X,T) + \epsilon
\end{align}
Where $\mu(X,T) = \EE[Y\mid X,T]$ so that $\EE[\epsilon\mid X,T] = 0$. Now the error of the pseudo-outcome from $\tau(X)$ is:
\begin{align}
    \frac{T-e(X)}{e(X)\circbrac{1-e(X)}}Y - \circbrac{\mu(X,1)-\mu(X,0)} &= \frac{\circbrac{T-e(X)}\circbrac{\mu(X,T) + \epsilon}}{e(X)\circbrac{1-e(X)}} - \circbrac{\mu(X,1)-\mu(X,0)} \\
    &= \begin{cases}
        \frac{\circbrac{\mu(X,1) + \epsilon}}{e(X)} - \circbrac{\mu(X,1)-\mu(X,0)} \text{ if $T=1$}\\
        -\frac{\circbrac{\mu(X,0) + \epsilon}}{1-e(X)} - \circbrac{\mu(X,1)-\mu(X,0)}\text{ if $T=0$}
    \end{cases} \\
    &= \begin{cases}
        \frac{1}{e(X)} \circbrac{ \circbrac{1-e(X)} \mu(X,1) + e(X)\mu(X,0) + \epsilon} \text{ if $T=1$}\\
        \frac{-1}{1-e(X)} \circbrac{{e(X)\mu(X,0) + \circbrac{1-e(X)}\mu(X,1) - \epsilon}} \text{ if $T=0$} \label{eq:final_line_pseudo_eq_1}
    \end{cases}
\end{align}
If we let $\phi(X) = \circbrac{1-e(X)} \mu(X,1) + e(X)\mu(X,0)$ then we have that \eqref{eq:final_line_pseudo_eq_1} is equal to:
\begin{align}
    =\frac{T-e(X)}{e(X)\circbrac{1-e(X)}}\circbrac{\phi(X) + \circbrac{-1}^{T+1}\epsilon}
\end{align}
Now if we let $\Tilde{\epsilon} = \circbrac{-1}^{T+1}\epsilon\frac{T-e(X)}{e(X)\circbrac{1-e(X)}}$ we can see that we now have $\EE\squarebrac{\Tilde{\epsilon}\mid X,T }=0$. Moreover, since the distribution of $\Tilde{\epsilon}$ given $X,T$ is just a constant scaled version of $\epsilon$ we have that $\Tilde{\epsilon}\mid X,T$ is Gaussian if and only if ${\epsilon}\mid X,T$ is.
\end{proof}

\subsection{Closed Form Posterior Expressions}

Now, for the closed form expressions have as follows, were the training dataset is $\{\bx_i,t_i,\Tilde{y}_i-\hat{\omega}_o(\bx_i)\}^{n_e}_{i=1}$:
\begin{align}
    \bM_N &=\circbrac{\circbrac{\bK(\bx_i,\bx_j)}_{t_i,t_j}}_{i,j} \\
    \boldsymbol{\Sigma}_N &= \mathrm{diag}\circbrac{\circbrac{\sigma^2_{t_i}}_i} \\
    \by_N &= \circbrac{\Tilde{y}_i-\hat{\omega}_o(\bx_i)}_i\\
    \bk_N(\bx) &= \circbrac{k(\bx_i,\bx)}_i \\
    \Tilde{\Delta}_{\cD_e}(\bx) &= \bk_N(\bx)^{\top} (\bM_N +\boldsymbol{\Sigma}_N)^{-1}\by_N \\
    k_{\cD_e}(\bx,\bx^{\prime}) &= k(\bx,\bx^{\prime}) - \bk_N(\bx)^{\top}(\bM_N +\boldsymbol{\Sigma}_N)^{-1}\bk_N(\bx^{\prime})
\end{align}
\subsection{Closed Form Bounds}
\begin{customthm}{\ref{thm:uniform_bounds}}
    Let the posterior for $\Delta(\cdot)$ from the GP model defined in Section \ref{sec:pseudo_GP} be given pointwise by $\cN(\Tilde{\Delta}(\bx),\sigma^{2}(\bx))$ where $\Tilde{\Delta},\sigma: \cX_o \to \mathbb{R}$. Then, under assumption \ref{assump:GP}, for fixed $\delta \in (0,1),\tau\in \mathbb{R}^{+}$ we have:
\begin{align*}
    P(\lvert\Delta(\bx) - \Tilde{\Delta}(\bx) \rvert& \leq B(\bx) \hspace{0.5em} \forall \bx \in \cX_o) > 1- \delta \\
    B(\bx) &= \sqrt{2\log \circbrac{\frac{M\circbrac{\tau,\cX_o}}{\delta}}}\sigma(\bx) \\
    &+ \gamma(\tau,\bX_e,L_k,L_{\Delta})
\end{align*}
Where $M\circbrac{\tau,\cX_o}$ is the $\tau$ covering number of $\cX_o$, defined as the minimum number of spherical balls of radius $\tau$ needed to cover $\cX_o$, and $\gamma(\tau,\bX_e,L_k,L_{\Delta})$ is defined in Appendix \ref{ap:guassian_process}.
\end{customthm}
\begin{proof}
    This follows from theorem 3.1 in \citet{lederer2019uniform} which we reproduce here:
\begin{customthm}{}[\citet{lederer2019uniform}]
 Consider a zero mean Gaussian process defined through the continuous covariance kernel $k(\cdot, \cdot)$ with Lipschitz constant $L_k$ on the compact set $\mathbb{X}$. Furthermore, consider a continuous unknown function $f: \mathbb{X} \rightarrow \mathbb{R}$ with Lipschitz constant $L_f$ and $N \in \mathbb{N}$ observations $y_i$ satisfying:
\begin{assumption}
    The unknown function $f(\cdot)$ is a sample from a Gaussian process $\mathcal{G} \mathcal{P}\left(0, k\left(\boldsymbol{x}, \boldsymbol{x}^{\prime}\right)\right)$ and observations $y=f(\boldsymbol{x})+\epsilon$ are perturbed by zero mean i.i.d. Gaussian noise $\epsilon$ with variance $\sigma_n^2$.
\end{assumption} 
 
Then, the posterior mean function $\nu_N(\cdot)$ and standard deviation $\sigma_N(\cdot)$ of a Gaussian process conditioned on the training data $\left\{\left(\boldsymbol{x}_i, y_i\right)\right\}_{i=1}^N$ are continuous with Lipschitz constant $L_{\nu_N}$ and modulus of continuity $\omega_{\sigma_N}(\cdot)$ on $\mathbb{X}$ such that:

$$
\begin{aligned}
L_{\nu_N} & \leq L_k \sqrt{N}\left\|\left(k\left(\boldsymbol{X}_N, \boldsymbol{X}_N\right)+\sigma_n^2 \boldsymbol{I}_N\right)^{-1} \boldsymbol{y}_N\right\| \\
\omega_{\sigma_N}(\tau) & \leq \sqrt{2 \tau L_k\left(1+N\left\|\left(k\left(\boldsymbol{X}_N, \boldsymbol{X}_N\right)+\sigma_n^2 \boldsymbol{I}_N\right)^{-1}\right\| \max _{\boldsymbol{x}, \boldsymbol{x}^{\prime} \in \mathbb{X}} k\left(\boldsymbol{x}, \boldsymbol{x}^{\prime}\right)\right)}
\end{aligned}
$$

Moreover, pick $\delta \in(0,1), \tau \in \mathbb{R}_{+}$and set

$$
\begin{aligned}
& \beta(\tau)=2 \log \left(\frac{M(\tau, \mathbb{X})}{\delta}\right) \\
& \gamma(\tau)=\left(L_{\nu_N}+L_f\right) \tau+\sqrt{\beta(\tau)} \omega_{\sigma_N}(\tau)
\end{aligned}
$$

Then, it holds that

$$
P\left(\left|f(\boldsymbol{x})-\nu_N(\boldsymbol{x})\right| \leq \sqrt{\beta(\tau)} \sigma_N(\boldsymbol{x})+\gamma(\tau), \forall \boldsymbol{x} \in \mathbb{X}\right) \geq 1-\delta
$$
\end{customthm}
In our case, the bounds on the Lipschitz constants and modulus of continuity are different. However by the same argument as \citet{lederer2019uniform} we have:
\begin{align}
    L_{\Tilde{\Delta}_{\cD_e}} &\leq L_k \sqrt{N}\lVert (\bM_N +\boldsymbol{\Sigma}_N)^{-1}\by_N \rVert\\ 
    \omega_{\sigma_N}(\tau) &\leq \sqrt{2 \tau L_k\left(1+N\left\|(\bM_N +\boldsymbol{\Sigma}_N)^{-1}\right\| \max _{\boldsymbol{x}, \boldsymbol{x}^{\prime} \in \mathbb{X}} k\left(\boldsymbol{x}, \boldsymbol{x}^{\prime}\right)\right)}
\end{align}
Putting this in to the above Theorem of \citet{lederer2019uniform} that, under assumption \ref{assump:GP}, for fixed $\delta \in (0,1),\tau\in \mathbb{R}^{+}$ we have
\begin{align*}
    P(\lvert\Delta(\bx) - \Tilde{\Delta}(\bx) \rvert& \leq B(\bx) \hspace{0.5em} \forall \bx \in \cX_o) > 1- \delta \\
    B(\bx) &= \sqrt{2\log \circbrac{\frac{M\circbrac{\tau,\cX_o}}{\delta}}}\sigma(\bx) \\
    &+ \left(L_k \sqrt{N}\lVert (\bM_N +\boldsymbol{\Sigma}_N)^{-1}\by_N \rVert+L_f\right) \tau\\
    &+\sqrt{\beta(\tau)} \sqrt{2 \tau L_k\left(1+N\left\|(\bM_N +\boldsymbol{\Sigma}_N)^{-1}\right\| \max _{\boldsymbol{x}, \boldsymbol{x}^{\prime} \in \mathbb{X}} k\left(\boldsymbol{x}, \boldsymbol{x}^{\prime}\right)\right)}
\end{align*}
Where we can see the claimed convergence property in $\tau$. 
\end{proof}
\subsection{LCM Kernel and Causal Multitask Kernel of \citet{alaa2017bayesian}}\label{ap:LCM_comparison}

In this the work of \citet{alaa2017bayesian}, CATE is modelled using a multitask Gaussian process \citep{bonilla2007multi}. Multitask Gaussian Processes use a GP in vector-valued Reproducing Kernel Hilbert Space (vv-RKHS) to share information between tasks \citep{alvarez2012kernels}. In \citet{alaa2017bayesian}, learning the conditional outcome function for each treatment is seen as a separate task, so we jointly model:
\begin{align}
    Y | \mathbf{x},t \sim \cN(0,f_t({\bx}),\sigma_t^2) 
\end{align}
Where each $f_t$ is a Gaussian Process. The kernel $ \Tilde{\bK}_{\eta}:\cX \times \cX \to \mathbb{R}^{2 \times 2}$ is now a symmetric
positive semi-definite matrix-valued function, with hyper-parameters $\eta$. In the case of \citet{alaa2017bayesian} they use a \emph{linear model of coregionalization}\footnote{See \citet{alvarez2012kernels} for more details.}, giving the kernel as:
\begin{align}
    \Tilde{\bK}(\bx,\bx^{\prime}) = \bA_0 k(\bx,\bx^{\prime}) + \bA_1 l(\bx,\bx^{\prime})
\end{align}
Where $\bA_t$ is given by:
\begin{align}
    \Tilde{\mathbf{A}}_0=\left[\begin{array}{cc}
\theta_{00}^2 & \rho_0 \\
\rho_0 & \theta_{01}^2
\end{array}\right], \Tilde{\mathbf{A}}_1=\left[\begin{array}{cc}
\theta_{10}^2 & \rho_1 \\
\rho_1 & \theta_{11}^2
\end{array}\right].
\end{align}
And $\bA_t$ are now free hyperparameters to learn. 

This is equivalent to the LCM kernel that we make use of for our experiments, however we regress onto pseudo-outcomes as opposed to observed outcomes. This is equivalent to scaling the task $t$ is scaled by $\frac{t-\pi(\bx))}{\pi(\bx)(1-\pi(\bx))}$ and leaving all hyper-parameters free to learn. As scaled Gaussian processes are still Gaussian processes this is same as using the kernel:
\begin{align}
    {\bK}(\bx,\bx^{\prime}) = \bM\bA_0 k(\bx,\bx^{\prime}) + \bM\bA_1 l(\bx,\bx^{\prime})
\end{align}
Where:
\begin{align}
    \bM = \circbrac{\circbrac{\frac{t-\pi(\bx))}{\pi(\bx)(1-\pi(\bx))}}\circbrac{\frac{t^{\prime}-\pi(\bx))}{\pi(\bx)(1-\pi(\bx))}}}_{t,t^{\prime}=0,1}
\end{align}
This demonstrates an equivalence between the Pseudo-outcome based LCM method we use for the experiments and the methods of \citep{alaa2017bayesian}.
\section{Experiment Details}\label{ap:experiment_details}

\subsection{Model Tuning Details}\label{ap:model_exp_details}
For each of the models we regress from $\bx,\bt$ onto $\Tilde{y}-\hat{w}_o(\bx)$ where $\hat{w}_o(\bx)$ is our estimate of ${w}_o(\bx)$. We do so as:
\begin{align}
    \Tilde{Y}-\hat{w}_o(X) \sim \cN(f_t(X),\sigma^2_t)
\end{align}
With specific model details as follows:
\begin{enumerate}
    \item \textbf{Standard or Naive GP}  Taking $f_0 = f_1=f $ and $\sigma^2_0 = \sigma^2_1=\sigma^22$. We then model $f$ directly using a $\mathrm{GP}(0,k_\theta)$ where $k_\theta$ is a kernel with hyper parameters $\theta$. Described throughout as standard or naive GP. The hyperparmeters are given by $\theta,\sigma$.
    \item \textbf{LCM GP}  Modelling $f_t$ using a multitask GP. This corresponds to using the vector valued kernel:
    \begin{align}
    {\bK}(\bx,\bx^{\prime}) = \bA_0 k_{\theta}(\bx,\bx^{\prime}) + \bA_1 l_{\eta}(\bx,\bx^{\prime})
\end{align}
Where $\bA_t$ is given by:
\begin{align}
    {\mathbf{A}}_0=\left[\begin{array}{cc}
\theta_{00}^2 & \rho_0 \\
\rho_0 & \theta_{01}^2
\end{array}\right], {\mathbf{A}}_1=\left[\begin{array}{cc}
\theta_{10}^2 & \rho_1 \\
\rho_1 & \theta_{11}^2
\end{array}\right].
\end{align}
CATE differences are then formed as the weighted average between both treatments. So modelled as:
\begin{align}
    \Delta(\bx) = \sum^1_{t=0}f_t(\bx)
\end{align}
Where $f_t$ is the prediction for task $t$. For this method the hyper-parameters to learn are $\theta,\eta,{\mathbf{A}}_0,{\mathbf{A}}_1,\sigma_0,\sigma_1$.
\item \textbf{Our Approach}. We use a multitask Gaussian process given by:
\begin{align}
    \bK &= \ba\ba^{\top} k_{\theta} + \bb\bb^{\top} l_{\eta}\label{eq:matrix_parameters}\\
        \ba &= \begin{bmatrix}
1 & 1 & 1\\
\end{bmatrix} \\
\bb &= \begin{bmatrix}
\frac{-1}{1-\pi(X)} & \frac{1}{\pi(X)} & 0\\
\end{bmatrix}
\end{align}
Where the first task models $f_0$, the second $f_1$, and the final is the CATE gap, $\Delta(\bx)$.
Using the decomposition:
\begin{align*}
    \Tilde{Y} = \tau(X) + \circbrac{\frac{T-\pi(X)}{\pi(X)\circbrac{1-\pi(X)}}} \phi(X) + \Tilde{\epsilon}
\end{align*}
This is equivalent to modelling $\tau(X) \sim \mathrm{GP}(0,k_\theta)$ and  $\phi(X) \sim \mathrm{GP}(0,l_\eta)$. The hyper-parameters for this method are $\eta,\theta,\sigma_1,\sigma_0$.
\end{enumerate}
\subsection{Simulation Details}\label{ap:simulation_details}
For the first experiment, we simulate data as follows
\begin{align}
    X| E=o \sim \cU([-3,3]^d)&, \hspace{0.5em}  X| E=e \sim \cU([-1,1]^d) \\
    Y_{|X=\bx,T=t, E=o} &= \sum^2_{i=0}\sum^1_{j=1}\beta_{i,j}^{\top}(t^j \odot \bx^j) + \epsilon \\
    Y_{|X=\bx,T=t, E=o} &= \sum^2_{i=0}\sum^1_{j=1}\beta_{i,j}^{\top}(t^j \odot \bx^j) +\bf_t(\bx) +\epsilon \\
    \epsilon \sim \cN(0,\sigma_0),& \bf_t \sim GP(0,k_{\theta_0})
\end{align}
Where $\beta_{i,j}=1$ for each $i,j$ and $\sigma_0 = 0.5$ 

\subsection{IHDP details}\label{ap:IHDP_details}
We simulate covariates following a similar approach to \citet{hussain2022falsification}. For the observational dataset we uniformly sample from the IHDP covariate distribution. For the experimental covariate distribution we sample with weights:
\begin{align}
    w_i = 0.8^{\mathbbm{1}\setbrac{\text{mother is smoker}} +\mathbbm{1}\setbrac{\text{is male}}}
\end{align}
So that the experiment dataset is significantly more likely to include male babies whose mothers are smokers. For the experimental dataset treatment is simulated as $\mathrm{Ber}(p)$. The outcome is then simulated as:
\begin{align}
        Y_{|X=\bx,T=t, E=o} &= \sum^1_{i=0}\sum^1_{j=1}\beta_{i,j}^{\top}(t^j \odot \bx^j) + \epsilon \\
    Y_{|X=\bx,T=t, E=o} &= \sum^1_{i=0}\sum^1_{j=1}\beta_{i,j}^{\top}(t^j \odot \bx^j) +\bf_t(\bx) +\epsilon \\
    \epsilon \sim \cN(0,\sigma_0),& \bf_t \sim GP(0,k_{\theta_0})
\end{align}
Where, $\beta_{i,j}= Z_i N_j$ where $Z_i \sim \mathrm{Ber}(0.3)$ and $N_j \sim \cN(0,1)$ and $\sigma_0 = 0.5$. 
\section{Additional Results}

\subsection{Simulated Experiment Additional Results}\label{ap:simulated_additional_results}

\begin{table}[H]
    \centering
     \resizebox{\linewidth}{!}{\begin{tabular}{ |c||c|c|c||c|c|c||c|c|c|  }
 \hline
  & \multicolumn{3}{|c|}{MSE}& \multicolumn{3}{|c|}{Coverage}& \multicolumn{3}{|c|}{Interval Width} \\
 \cline{2-10}
 Dim $X$ & Ours & Standard & LCM & Ours & Standard & LCM & Ours & Standard & LCM \\
 \hline
 $5$   & $0.301\pm0.006$    & $1.21\pm0.06$ &   $0.352\pm0.007$& $0.935\pm0.003$    & $0.858\pm0.008$ &   $0.932\pm0.003$ & $1.97 \pm0.02$    & $3.22 \pm0.02$ &   $2.15 \pm0.02$\\
 \hline
 $10$&   $1.77 +\pm 0.0157$  & $2.05 \pm 0.02$   & $1.91 \pm0.02$    & $0.785 +\pm0.005$  & $0.796 \pm 0.007$   & $0.303 \pm0.021$    &$3.31\pm0.03$&   $3.65 \pm 0.04$&   $1.09 \pm 0.08$\\
 \hline
 $25$ & $2.15 \pm 0.02$ & $2.27\pm0.03$& $2.02\pm0.01$ & $0.779 \pm0.004$   & $0.754 \pm 0.008$&   $0.128 \pm 0.012$& $3.60 \pm 0.02$   & $3.50 \pm 0.04$&   $0.463\pm 0.045$\\
 
 \hline
\end{tabular}}
    \caption{In distribution results for $n_{\exp}=1000$ across dimension, average across 200 runs with 95\% confidence interval.}
\end{table}

\begin{table}[H]
    \centering
     \resizebox{\linewidth}{!}{\begin{tabular}{ |c||c|c|c||c|c|c||c|c|c|  }
 \hline
  & \multicolumn{3}{|c|}{MSE}& \multicolumn{3}{|c|}{Coverage}& \multicolumn{3}{|c|}{Interval Width} \\
 \cline{2-10}
 Dim $X$ & Ours & Standard & LCM & Ours & Standard & LCM & Ours & Standard & LCM \\
 \hline
 $5$   & $2.10\pm0.05$    & $2.17\pm0.06$ &    $2.31\pm0.07$ & $0.825\pm0.010$    & $0.817\pm0.010$ &   $0.997\pm0.01$ & $3.91 \pm0.00$    & $3.92 \pm0.00$ &   $9.65 \pm 0.20$\\
 \hline
 $10$&   $2.01 \pm 0.02$  & $2.05 \pm 0.02$   & $2.38 \pm0.02$& $0.832 \pm 0.002$    &$0.829 \pm0.003 $&   $0.720 \pm 0.041$& $3.91 \pm 0.02$    & $3.91 \pm 0.02$&   $3.85 \pm 0.15$\\
 \hline
 $25$ &$2.01 \pm 0.01$ & $2.03\pm0.01$& $2.04\pm0.00$& $0.832 \pm 0.001$   & $0.831 \pm 0.002$&   $0.239\pm 0.030$   & $3.92 \pm 0.00$&   $3.92 \pm 0.00$ & $0.910 \pm 0.14$\\
 \hline
\end{tabular}}
    \caption{Out of distribution results for $n_{\exp}=1000$ across dimension, average across 200 runs with 95\% confidence interval. }
\end{table}

\begin{table}[H]
    \centering
     \resizebox{\linewidth}{!}{\begin{tabular}{ |c||c|c|c||c|c|c||c|c|c|  }
 \hline
  & \multicolumn{3}{|c|}{MSE}& \multicolumn{3}{|c|}{Coverage}& \multicolumn{3}{|c|}{Interval Width} \\
 \cline{2-10}
 Dim $X$ & Ours & Standard & LCM & Ours & Standard & LCM & Ours & Standard & LCM \\
 \hline
 $5$   & $0.155\pm0.002$    & $0.808\pm0.06$ &   $0.290 \pm 0.010$& $0.950\pm0.002$    & $0.897\pm0.006$ &   $0.892\pm0.005$ & $1.48 \pm0.01$    & $2.86 \pm0.08$ &   $1.76 \pm0.03$\\
 \hline
 $10$&   $1.49 +\pm 0.01$  & $1.94 \pm 0.02$   & $1.69 \pm0.02$    & $0.807 +\pm0.003$  & $0.812 \pm 0.006$   & $0.481 \pm0.019$    &$3.17\pm0.02$&   $3.67 \pm 0.04$&   $1.70 \pm 0.07$\\
 \hline
 $25$ & $2.08 \pm 0.01$ & $2.19\pm0.02$& $2.01\pm0.02$ & $0.804 \pm0.002$   & $0.775 \pm 0.005$&   $0.104 \pm 0.008$& $3.742 \pm 0.01$   & $3.587 \pm 0.03$&   $0.373\pm 0.03$\\
 
 \hline
\end{tabular}}
    \caption{In distribution results for $n_{\exp}=2500$ across dimension, average across 200 runs with 95\% confidence interval.}
\end{table}

\begin{table}[H]
    \centering
     \resizebox{\linewidth}{!}{\begin{tabular}{ |c||c|c|c||c|c|c||c|c|c|  }
 \hline
  & \multicolumn{3}{|c|}{MSE}& \multicolumn{3}{|c|}{Coverage}& \multicolumn{3}{|c|}{Interval Width} \\
 \cline{2-10}
 Dim $X$ & Ours & Standard & LCM & Ours & Standard & LCM & Ours & Standard & LCM \\
 \hline
 $5$   & $2.13\pm0.11$    & $2.21\pm0.12$ &    $2.76\pm0.27$ & $0.819\pm0.011$    & $0.810\pm0.013$ &   $0.999\pm0.001$ & $3.92 \pm0.00$    & $3.92 \pm0.00$ &   $11.42 \pm 0.32$\\
 \hline
 $10$&   $2.01 \pm 0.02$  & $2.03 \pm 0.02$   & $2.33 \pm0.02$& $0.833 \pm 0.002$    &$0.831 \pm0.003 $&   $0.931 \pm 0.011$& $3.92 \pm 0.00$    & $3.92 \pm 0.00$&   $5.80 \pm 0.18$\\
 \hline
 $25$ &$1.99 \pm 0.01$ & $2.01\pm0.02$& $2.01\pm0.00$& $0.834\pm 0.001$   & $0.832 \pm 0.002$&   $0.216\pm 0.011$   & $3.92 \pm 0.00$&   $3.92 \pm 0.00$ & $0.794 \pm 0.18$\\
 \hline
\end{tabular}}
    \caption{Out of distribution results for $n_{\exp}=2500$ across dimension, average across 200 runs with 95\% confidence interval. }
\end{table}

\subsection{IHDP Experiment Additional Results}\label{ap:idhp_additional_results}

\begin{table}[H]
    \centering
     \resizebox{0.5\linewidth}{!}{\begin{tabular}{|c|c|c|c|}
    \hline
    \textbf{Model} &\textbf{MSE}  &\textbf{Coverage} & \textbf{Interval Width}\\
     \hline
        Ours & $\mathbf{1.19 \pm 0.04}$ & $\mathbf{0.795 \pm 0.008}$  & $\mathbf{2.572 \pm 0.02}$ \\
         \hline
        Naive GP & ${2.43 \pm 0.13}$  & ${0.752 \pm 0.010}$  & $3.42 \pm 0.03$ \\
         \hline
        LCM & ${1.57 \pm 0.05}$ & ${0.752 \pm 0.010}$ & ${3.21 \pm 0.05}$ \\
         \hline
    \end{tabular}}
    \caption{In of distribution results for the IHDP setting described in Section \ref{sec:semi_synthetic} with $n_{e} = 400$ averaged over 100 runs where now the experimental treatment proportion is $0.7$.}
\label{tab:simulated_results_main}
\end{table}

\begin{table}[H]
    \centering
     \resizebox{0.5\linewidth}{!}{\begin{tabular}{|c|c|c|c|}
    \hline
    \textbf{Model} &\textbf{MSE}  &\textbf{Coverage} & \textbf{Interval Width}\\
     \hline
        Ours & $\mathbf{1.93 \pm 0.04}$ & $\mathbf{0.812 \pm 0.008}$  & $\mathbf{3.65 \pm 0.02}$ \\
         \hline
        Naive GP & ${2.27 \pm 0.13}$  & ${0.797 \pm 0.010}$  & $3.81 \pm 0.03$ \\
         \hline
        LCM & ${2.50 \pm 0.05}$ & ${0.961 \pm 0.010}$ & ${8.03 \pm 0.05}$ \\
         \hline
    \end{tabular}}
    \caption{Out of distribution results for the IHDP setting described in Section \ref{sec:semi_synthetic} with $n_{e} = 400$ averaged over 100 runs where now the experimental treatment proportion is $0.7$.}
\label{tab:simulated_results_main}
\end{table}

\begin{table}[H]
    \centering
     \resizebox{0.5\linewidth}{!}{\begin{tabular}{|c|c|c|c|}
    \hline
    \textbf{Model} &\textbf{MSE}  &\textbf{Coverage} & \textbf{Interval Width}\\
     \hline
        Ours & $\mathbf{1.63 \pm 0.04}$ & $\mathbf{0.643 \pm 0.008}$  & $\mathbf{2.27\pm 0.02}$ \\
         \hline
        Naive GP & ${2.59 \pm 0.13}$  & ${0.732 \pm 0.010}$  & $3.48 \pm 0.03$ \\
         \hline
        LCM & ${2.04 \pm 0.05}$ & ${0.931 \pm 0.010}$ & ${6.33 \pm 0.05}$ \\
         \hline
    \end{tabular}}
    \caption{Out of distribution results for the IHDP setting described in Section \ref{sec:semi_synthetic} with $n_{e} = 400$ averaged over 100 runs where now the experimental treatment proportion is $0.7$.}
\label{tab:simulated_results_main}
\end{table}

\begin{table}[H]
    \centering
     \resizebox{0.5\linewidth}{!}{\begin{tabular}{|c|c|c|c|}
    \hline
    \textbf{Model} &\textbf{MSE}  &\textbf{Coverage} & \textbf{Interval Width}\\
     \hline
        Ours & $\mathbf{2.07 \pm 0.04}$ & $\mathbf{0.778 \pm 0.008}$  & $\mathbf{3.58 \pm 0.02}$ \\
         \hline
        Naive GP & ${2.37 \pm 0.13}$  & ${0.789 \pm 0.010}$  & $3.82 \pm 0.03$ \\
         \hline
        LCM & ${2.58 \pm 0.05}$ & ${0.931 \pm 0.010}$ & ${6.33 \pm 0.05}$ \\
         \hline
    \end{tabular}}
    \caption{Out of distribution results for the IHDP setting described in Section \ref{sec:semi_synthetic} with $n_{e} = 400$ averaged over 100 runs where now the experimental treatment proportion is $0.7$.}
\label{tab:simulated_results_main}
\end{table}

\subsection{Robustness Results}\label{ap:robustness_results}

We now repeat the experiment for the IHDP dataset but we add squared terms to the simulation as follows:

\begin{align}
        Y_{|X=\bx,T=t, E=o} &= \sum^1_{i=0}\sum^1_{j=1}\beta_{i,j}^{\top}(t^j \odot \bx^j) + \gamma_{i,j}^{\top}(t^j \odot \bx^j)^2 + \epsilon \\
    Y_{|X=\bx,T=t, E=o} &= \sum^1_{i=0}\sum^1_{j=1}\beta_{i,j}^{\top}(t^j \odot \bx^j) + \gamma_{i,j}^{\top}(t^j \odot \bx^j)^2+\bf_t(\bx) +\epsilon \\
    \epsilon \sim \cN(0,\sigma_0),& \bf_t \sim GP(0,k_{\theta_0})
\end{align}

We still fit a linear model for $\omega_{o}(\bx)$ which ensures that $\Delta(\bx)$ is not a GP. 

\begin{table}[H]
    \centering
     \resizebox{0.5\linewidth}{!}{\begin{tabular}{|c|c|c|c|}
    \hline
    \textbf{Model} &\textbf{MSE}  &\textbf{Coverage} & \textbf{Interval Width}\\
     \hline
        Ours & $\mathbf{1.10 \pm 0.03}$ & $\mathbf{0.824 \pm 0.03}$  & $\mathbf{2.63 \pm 0.02}$ \\
         \hline
        Naive GP & ${2.13 \pm 0.10}$  & ${0.761 \pm 0.01}$  & $3.39 \pm 0.04$ \\
         \hline
        LCM & ${1.34 \pm 0.04}$ & ${0.832 \pm 0.01}$ & ${2.96 \pm 0.04}$ \\
         \hline
    \end{tabular}}
    \caption{In of distribution results for the IHDP setting described in Section \ref{sec:semi_synthetic} with $n_{e} = 400$ averaged over 100 runs where now the experimental treatment proportion is $0.7$.}
\label{tab:simulated_results_main}
\end{table}

\begin{table}[H]
    \centering
     \resizebox{0.5\linewidth}{!}{\begin{tabular}{|c|c|c|c|}
    \hline
    \textbf{Model} &\textbf{MSE}  &\textbf{Coverage} & \textbf{Interval Width}\\
     \hline
        Ours & $\mathbf{1.91 \pm 0.03}$ & $\mathbf{0.821 \pm 0.003}$  & $\mathbf{3.66 \pm 0.05}$ \\
         \hline
        Naive GP & ${2.15 \pm 0.04}$  & ${0.805 \pm 0.004}$  & $3.80 \pm 0.014$ \\
         \hline
        LCM & ${2.22 \pm 0.09}$ & ${0.832 \pm 0.01}$ & ${7.76 \pm 0.27}$ \\
         \hline
    \end{tabular}}
    \caption{In of distribution results for the IHDP setting described in Section \ref{sec:semi_synthetic} with $n_{e} = 400$ averaged over 100 runs where now the experimental treatment proportion is $0.7$.}
\label{tab:simulated_results_main}
\end{table}

\section{Uniform Error Bounds}\label{ap:uniform_error_bounds}

Finally, we repeat the experiment in Section \ref{ap:robustness_results} but with the uniform error bounds. 
\begin{table}[H]
    \centering
    \begin{tabular}{|c|c|c|c|c|c|c|c|c|}
        \hline
         $n_e$ & $500$  & $1000$ & $5000$ & $10000$   \\
         \hline
        Whole Function Coverage &  $1.00 \pm 0.00 $& $1.00 \pm 0.00$  &$1.00 \pm 0.00$  &  $1.00 \pm 0.00$   \\
         \hline
        Interval width in Distribution &$21.3 \pm 0.12$  & $14.0 \pm 0.01$ & $5.36 \pm 0.01$ &   $3.76 \pm 0.03$  \\
         \hline
        Interval width out of distribution & $31.7 \pm 0.04$ & $30.4 \pm 0.04$ & $28.9 \pm 0.05$ & $28.6 \pm 0.1$    \\
         \hline
    \end{tabular}
    \caption{Uniform error bounds averaged over 100 runs. We vary the sample size to show that the in distribution bounds decrease in width as $n_e$ increases. }
    \label{tab:my_label}
\end{table}
\end{document}